\documentclass[11pt]{article}

\usepackage[final]{acl}

\usepackage{times}
\usepackage{latexsym}
\usepackage{multirow}
\usepackage{amsmath}
\usepackage[T1]{fontenc}

\usepackage[utf8]{inputenc}

\usepackage{microtype}

\usepackage{inconsolata}

\usepackage{graphicx}

\usepackage{booktabs}
\usepackage{xcolor}
\usepackage{float}
\usepackage{tabularx}

\definecolor{mygreen}{HTML}{e1f0d9}
\definecolor{myred}{HTML}{f7cdd0}
\definecolor{myblue}{HTML}{4371c4}
\newcommand{\greenbg}[1]{\colorbox{mygreen}{#1}}
\newcommand{\redbg}[1]{\colorbox{myred}{#1}}

\title{KoCo: Conditioning Language Model Pre-training on Knowledge Coordinates}

\author{Yudong Li\textsuperscript{1,2$\dagger$}, Jiawei Cai \textsuperscript{1$\dagger$}, Linlin Shen \textsuperscript{3,4*} \\
  \textsuperscript{1} School of Computer Science and Software Engineering, Shenzhen University \\
  \textsuperscript{2} Department of Electrical Engineering, Tsinghua University \\
  \textsuperscript{3} School of Artificial Intelligence, Shenzhen University \\
  \textsuperscript{4} Guangdong Provincial Key Laboratory of Intelligent Information Processing, Shenzhen University
}

\begin{document}
\maketitle
\begin{abstract}
  Standard Large Language Model (LLM) pre-training typically treats corpora as flattened token sequences, often overlooking the real-world context that humans naturally rely on to contextualize information. To bridge this gap, we introduce \textbf{Knowledge Coordinate Conditioning (KoCo)}, a simple method that maps every document into a three-dimensional semantic coordinate.  By prepending these coordinates as textual prefixes for pre-training, we aim to equip the model with explicit contextual awareness to learn the documents within the real-world knowledge structure. Experiment results demonstrate that KoCo significantly enhances performance across 10 downstream tasks and accelerates pre-training convergence by approximately 30\%. Furthermore, our analysis indicates that explicitly modeling knowledge coordinates helps the model distinguish stable facts from noise, effectively mitigating hallucination in generated outputs.
\end{abstract}

\section{Introduction}

\let\thefootnote\relax\footnotetext{$\dagger$ Equal Contribution\\*Corresponding Author: llshen@szu.edu.com}

The predominant paradigm in Large Language Model (LLM) pre-training treats corpora as flattened sequences of tokens. Models optimize the negative log-likelihood of the next token indiscriminately, whether that token originates from a peer-reviewed theorem or a fleeting dialogue on a social forum. While scaling laws have driven remarkable capabilities under this paradigm, the inherent heterogeneity of pre-training data remains a critical challenge. Humans do not learn from an undifferentiated stream of information; rather, we continually contextualize what we read based on its source and role within our knowledge system.

\begin{figure}[t]
  \centering
  \includegraphics[width=\linewidth]{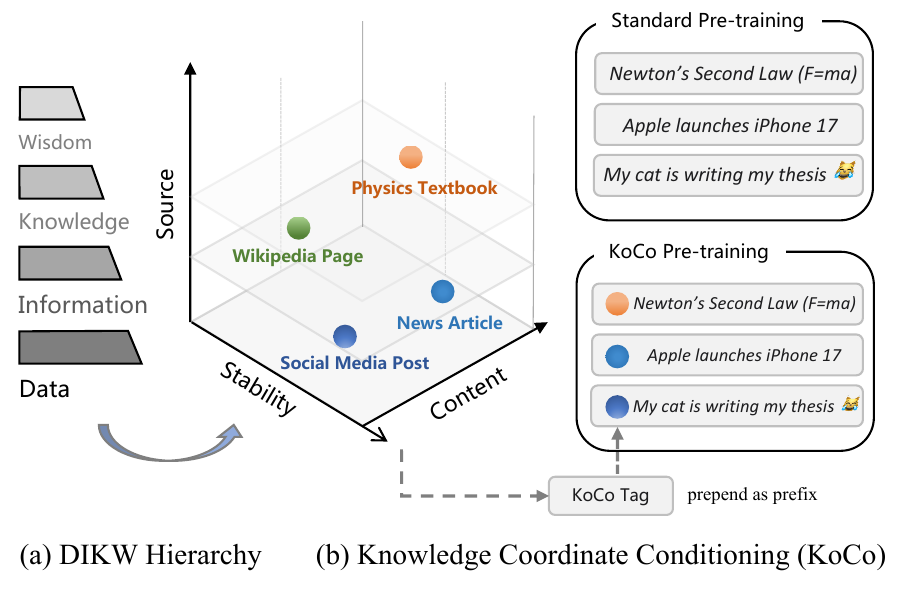}
  \caption{KoCo transforms DIKW into an objective representation of the knowledge contained in the corpus.}
  \label{fig:intro}
\end{figure}

To address this limitation, recent researches have explored different directions.
Metadata-aware pre-training explicitly informs the model of the data's origin. For instance, \citet{khalifa2024source} introduce source identifiers to enable knowledge attribution, while MeCo \cite{gaometadata} demonstrates that prepending source URLs can substantially accelerate convergence and improve performance.
The data selection approach aims to optimize the training distribution by prioritizing high-quality samples. Methods such as Ask-LLM \cite{sachdeva2024train} and other classifier-based filters \cite{wang2025ultra, li2024datacomp} have shown that training on carefully selected dataset subsets often outperforms training on the full dataset.

Building on these advancements, we hypothesize that efficient knowledge acquisition requires not just identifying the source or filtering by quality, but understanding the \textit{coordinate} of information within a broader knowledge space.
Existing source identifiers (i.e., URLs) are often too granular, rely on a priori mappings, and lack objectivity. 
For approximating human-like knowledge acquisition, where learners explicitly perceive the origin and role of information, a more detailed form of context is needed.
Meanwhile, data selection methods typically adopt a binary approach, which retains high-quality data and discards the rest. This is fundamentally different from human learning: when humans encounter lower-quality or noisy information, we do not simply erase it from our memory; instead, we contextualize it based on its origin and nature.

To bridge this gap, we introduce \textbf{Knowledge Coordinate Conditioning (KoCo)}, a method designed to provide the contextual grounding that has been notably absent in standard LLM pre-training. 
As shown in Figure~\ref{fig:intro}, our approach is inspired by the DIKW (Data-Information-Knowledge-Wisdom) hierarchy~\cite{rowley2007wisdom}, which suggests that raw data evolves into valuable knowledge only when it is contextualized and structurally connected. This mirrors the human cognitive process of organizing isolated information signals into a coherent understanding of the world. KoCo operationalizes this abstract concept by mapping every document into a concrete, three-dimensional semantic space defined by its Source (origin), Content (functional role), and Stability (temporal persistence). These dimensions function as an objective description, allowing the model to explicitly perceive meta-information of the corpus. For instance, it allows the model to distinguish between ``Evergreen Physics Theorem'' and ``Ephemeral Social Opinion'' during pre-training.
By acting as a cognitive abstraction layer, KoCo enables the model to learn knowledge representations at a higher level, facilitating a more robust understanding of the intricate relationships between diverse knowledge sources.
Our contributions are summarized as follows:

\begin{itemize}
    \item We introduce KoCo, a simple yet effective pre-training method. Experiments on 1.6B models show that KoCo improves downstream task performance, outperforming existing URL-based and data selection methods.
    \item We demonstrate that KoCo significantly improves training efficiency. When pre-training models from scratch, KoCo accelerates convergence by approximately 30\%.
    \item We show that KoCo effectively mitigates hallucination through conditional inference. By conditioning on reliable source coordinates, KoCo achieves a significant improvement on TruthfulQA.
    \item Our ablation studies reveal that each dimension of the coordinate system contributes complementarily to the performance, and the gains stem from the coordinate conditioning mechanism itself rather than knowledge distillation.
\end{itemize}

\begin{figure*}[!t]
    \centering
    
    \includegraphics[width=\linewidth]{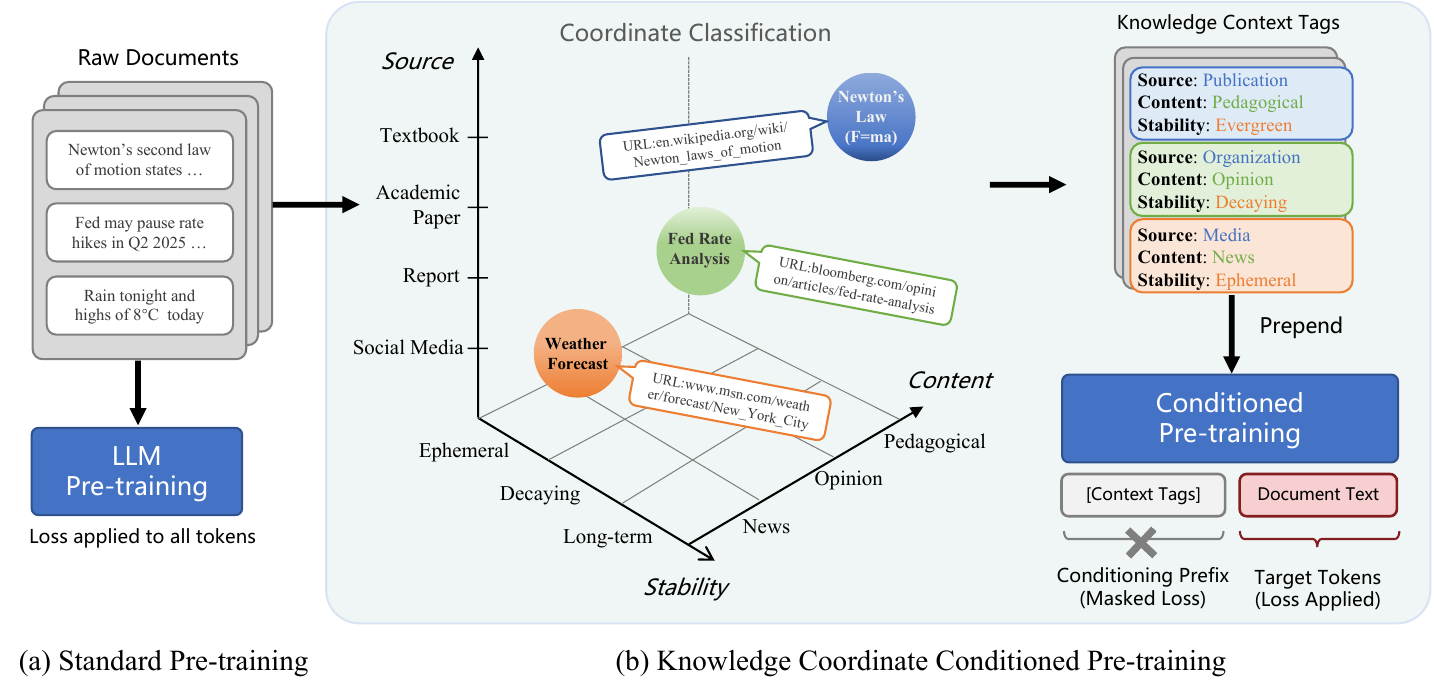}
    \caption{Overview of Knowledge Coordinate Conditioning (KoCo). Different from standard pre-training (a), KoCo (b) conditions the model on the context tag. By prepending it and applying masked loss, KoCo enables the model to distinguish knowledge properties.}
    \label{fig:overview}
    
\end{figure*}

\section{Related Work}

\subsection{Data Efficient Pre-training }

Scaling laws~\cite{kaplan2020scaling, hoffmann2022training} have established the fundamental relationships between model size, training data, and performance, demonstrating that larger models and more data lead to predictable improvements. However, efficiently utilizing pre-training data to train better models within limited computational budgets remains a critical challenge.

Recent work has explored various paradigms to enhance data efficiency during pre-training. 
Curriculum learning approaches schedule training data to present easier examples before harder ones, enabling models to learn more effectively~\cite{bengio2009curriculum, xu2020curriculum}. However, these methods rely on unified difficulty scoring and reordering mechanisms, which are challenging to implement at web scale. Data combination strategies~\cite{xie2023doremi, shen2023slimpajama, gadre2023datacomp} have explored the impact of domain-level data mixing on performance, demonstrating that appropriate combinations of the same data sources can substantially improve pre-training effectiveness. These approaches optimize the sampling proportions across different domains to achieve better downstream task performance.

More recently, metadata-aware pre-training has emerged as a promising direction. MeCo~\cite{gaometadata} introduces a condition and then cool-down strategy, achieving acceleration in pre-training convergence. 
Our work extends this line of research by introducing a Knowledge Coordinate. It equips the model with explicit contextual awareness regarding the nature of information.

\subsection{Data Selection and Synthesis}

Data selection methods aim to improve pre-training efficiency by curating higher-quality training corpora.
The mainstream approach employs classifier-based filtering to select high-quality data subsets. For instance, DataComp-LM~\cite{li2024datacomp} trains fastText classifiers to distinguish educational content from general web text, while ASK-LLM~\cite{sachdeva2024train} leverages instruction-tuned LLMs to directly evaluate sample quality, demonstrating that carefully selected subsets can match or exceed full-dataset performance with significantly reduced training time.
Alternative selection paradigms have also been explored, including diversity-based sampling and deduplication~\cite{sorscher2022beyond}, perplexity-based filtering~\cite{marion2023less}, and influence function approaches~\cite{xia2024less} that estimate the downstream impact of individual training samples.

To address the growing demand for high-quality training data, data synthesis has emerged as an important direction. 
\citet{gunasekar2023textbooks} demonstrates that models trained on synthetically generated textbook-style data can achieve strong reasoning capabilities, while \citet{kang2025demystifying} reveal that mixing approximately one-third of rephrased synthetic data with natural web text can effectively accelerate pre-training.
These approaches offer promising ways to augment limited high-quality data sources.

Our method is complementary to these approaches: whereas data selection and synthesis aim to \textit{refine} the training distribution by filtering or rewriting, KoCo focuses on \textit{contextualizing} the data. This allows the model to effectively utilize a broader range of corpora by explicitly recognizing their attributes, rather than discarding them.
\section{Approach}

Our goal is to transform the standard pre-training objective from modeling a flattened token sequence $P(x)$ to modeling a context-conditioned distribution $P(x|\mathcal{T})$, where $\mathcal{T}$ represents the knowledge coordinate of the document. 
This formulation aims to impose a context-aware knowledge acquisition process on LLM pre-training, enabling the model to explicitly perceive the position of each document within the global context of world knowledge. Formally, the KoCo training objective is:
\begin{equation}
\mathcal{L}_{\text{KoCo}} = -\sum_{i=1}^{n} \log P_\theta(x_i | x_{<i}, \mathcal{T})
\end{equation}
where $x = (x_1, \ldots, x_n)$ denotes the document tokens and $\mathcal{T}$ is the knowledge coordinate. 

Specifically, Knowledge Coordinate is a method that maps heterogeneous web data into a disentangled semantic space.
As shown in Figure~\ref{fig:overview}, we define the knowledge coordinate as a triplet $\mathcal{T} = (s, c, t)$, corresponding to Source, Content, and Stability. These dimensions are agnostic to specific semantic topics (e.g., physics or math).
Instead of providing supplementary factual knowledge, they serve as coordinates that simulate the contextual awareness humans naturally possess, which has been absent in standard model pre-training:

\begin{itemize}
   \item \textbf{Source} describes the origin, channel, and carrier of the information (e.g., academic journal vs. social media comment).

   \item \textbf{Content} characterizes the format and linguistic style of the text's expression (e.g., a pedagogical tutorial vs. a raw data log).

   \item \textbf{Stability} reflects the temporal persistence and expected validity period, inferred from the information's update frequency (e.g., a fundamental theorem is ``Evergreen'', while a weather report is ``Ephemeral'').
\end{itemize}

Table~\ref{tab:taxonomy} details the label set for each dimension. Evaluations on the DCLM \cite{li2024datacomp} corpus show that over 99.5\% of documents can be successfully mapped to this coordinate system, confirming its coverage of web-scale data.
For scalable implementation, we employ a lightweight language model Qwen-3-4B to act as a tagger. It takes the source URL and document text as input to predict the label for each dimension. These labels are concatenated to form a natural language prefix (e.g., \textit{Source: Academic; Content: Reference; Stability: Evergreen}) and prepended to the raw text. Only the loss over the document tokens is calculated during pre-training. This provides the model with context information when learning documents, mirroring the human learning process.

\textbf{Comparison with data-selection/distillation methods.} 
Our design differs fundamentally from existing approaches in data selection and distillation. 
Data selection methods rely on scalar quality scores to perform binary filtering (keep vs. drop), implicitly assuming that low-quality data contributes little to learning.
Data distillation methods seek to rewrite or synthesize content to improve quality, which is computationally expensive and risks losing the diversity of the original distribution.
In contrast, KoCo treats data properties as a multi-dimensional coordinate system rather than a single quality axis. By conditioning on these coordinates, KoCo allows the model to learn from diverse data distributions (including noisy or informal text), while explicitly understanding their limitations.

\begin{table}[t]
\centering
\small
\begin{tabular}{l|p{5.5cm}}
\toprule
\textbf{Dimension} & \textbf{Labels and Examples} \\
\midrule
\multirow{4}{*}{\textbf{Source}} 
& \textbf{Academic}, \textbf{Publication} (Books/Journals), \\
& \textbf{Government}, \textbf{Organization}, \textbf{Industry}, \\
& \textbf{Media}, \textbf{Community}, \\
& \textbf{Personal} (Blogs), \textbf{Codebase}, \textbf{Others} \\
\midrule
\multirow{5}{*}{\textbf{Content}} 
& \textbf{Instruction}, \textbf{Pedagogical} (Textbook), \\
& \textbf{Reference}, \textbf{Data} (Logs/Tables), \\
& \textbf{News}, \textbf{Opinion}, \textbf{Discussion}, \textbf{Review}, \\
& \textbf{Narrative-Nonfiction}, \textbf{Narrative-Fiction}, \\
& \textbf{Others} \\
\midrule
\multirow{4}{*}{\textbf{Stability}} 
& \textbf{Ephemeral} (e.g., stock prices), \\
& \textbf{Decaying} (e.g., software docs), \\
& \textbf{Long-term} (e.g., history), \\
& \textbf{Evergreen} (e.g., math, logic) \\
\bottomrule
\end{tabular}
\caption{\label{tab:taxonomy} The KoCo Taxonomy. This classification scheme categorizes documents based on their objective nature rather than their semantic topic.}
\end{table}

\begin{table*}[ht]
  \centering
  \resizebox{\linewidth}{!}{
      \begin{tabular}{lccccccccccc}
        \toprule
        \textbf{Method} & \textbf{COPA} & \textbf{ARC-e} & \textbf{ARC-c} & \textbf{CSQA} & \textbf{IFEval} &
        \textbf{OBQA}  & \textbf{PIQA} & \textbf{SIQA} & \textbf{LogiQA}    & \textbf{TruQA} & \textbf{Avg} \\
        \midrule
        URL Prefix (MeCo) & 82.00 & 75.40 & 44.37 & 63.96 & 20.02 & 50.80 & 73.00 & 52.90 & 25.50 & 36.33 & 52.42 \\

        \textit{+ Standard CPT} & 82.00 & 74.60 & 42.83 & 59.54 & 22.18 & 49.60 & 72.90 & 52.70 & 24.88 & 35.24 & 51.64 \\
        \textit{+ Data Selection} & 82.00 & 75.00 & 44.62 & 63.31 & 22.42 & 49.00 & 74.00 & 52.60 & 25.19 & 35.53 & 52.36 \\
        \midrule
        KoCo & 83.00 & 77.40 & 44.11 & 61.83 & 25.54 & 51.20 & 74.80 & 53.40 & 26.88 & 36.61 & 53.48 \\
        &
         \greenbg{$\uparrow 1.00$} & \greenbg{$\uparrow 2.00$} & \redbg{$\downarrow 0.26$} & \redbg{$\downarrow 2.13$} & \greenbg{$\uparrow 5.52$} & \greenbg{$\uparrow 0.40$} & \greenbg{$\uparrow 1.80$} & \greenbg{$\uparrow 0.50$} & \greenbg{$\uparrow 1.38$} & \greenbg{$\uparrow 0.28$} & \greenbg{$\uparrow 1.06$} \\
        \bottomrule
      \end{tabular}
  }
  \caption{Experiment results of pre-training a 1.6B language model on 8B tokens from DCLM. KoCo outperforms existing methods on most tasks and achieves a higher average score.}
  \label{tab:main_experiments}
\end{table*}
\section{Experiments}

\subsection{Experiment Setup}

\textbf{Model Structure and Initialization.} We use LLaMA \cite{dubey2024llama} architecture with LLaMA-3 tokenizer in all our experiments. To ensure a fair and controlled comparison, we conduct our main experiments using MeCo \cite{gaometadata} 1.6B checkpoint for initialization then compare the performance of different training paradigms. Following MeCo, we adopt DCLM as our corpus, which allows for a fair comparison. We randomly select a 100GB disk size subset for training. We also conduct pre-training from scratch using identical structure and data with 0.6B model parameters. The experiments are conducted using 4*A800 80G GPUs.

\textbf{Evaluation.} We adopt the OLMES suite \cite{gu2025olmes} for evaluation, which includes the following tasks: COPA~\cite{roemmele2011choice}, ARC-Easy (ARC-e)~\cite{clark2018think}, ARC-Challenge (ARC-c)~\cite{clark2018think}, CommonsenseQA (CSQA)~\cite{talmor2019commonsenseqa}, IFEval~\cite{zhou2023instruction}, OpenBookQA (OBQA)~\cite{mihaylov2018can}, PIQA~\cite{bisk2020piqa}, Social IQA (SIQA)~\cite{sap2019socialiqa}, LogiQA~\cite{liu2020logiqa}, and TruthfulQA (TruQA)~\cite{lin2022truthfulqa}. All results correspond to 5-shot in-context learning unless explicitly noted otherwise. We evaluate all samples in each task to ensure comprehensive and reliable results. During inference, all models use the same prompt for inference (without adding any prefixes).

\subsection{KoCo Improves Model Performance on Downstream Tasks}

For experiment, we conduct continual pre-training on MeCo 1.6B checkpoint, which is a strong baseline using URL as conditioning. We compare our KoCo with the following settings:

\begin{itemize}
   \item \textbf{URL Prefix Baseline (MeCo)}: The original MeCo checkpoint released by \citet{gaometadata}, which uses URL as a conditioning signal and achieves 33\% of pre-training acceleration.
    \item \textbf{Standard CPT}: Continual pre-training on the MeCo checkpoint using standard training paradigm with a randomly selected subset of 100GB DCLM corpus.
    \item \textbf{Data Selection}: Continual pre-training using a higher-quality subset (and same data size) selected by the fastText classifier \cite{li2024datacomp}, retaining the top 50\% of documents.
    \item \textbf{KoCo (ours)}: Continual pre-training using our proposed knowledge coordinate conditioning with a randomly selected subset of DCLM identical to the standard CPT setting.
\end{itemize}

As shown in Table~\ref{tab:main_experiments}, the experimental results reveal several findings. First, standard continual pre-training actually degrades the performance of the MeCo checkpoint, which is consistent with findings in prior work and confirms the strength of MeCo as the baseline. Second, the data selection approach achieves comparable performance to MeCo, but the limited amount of high-quality data makes it difficult to substantially improve model performance. Most importantly, KoCo significantly outperforms all baselines. Notably, KoCo achieves this improvement using the same data source as MeCo (DCLM corpus), without introducing any external data unseen by MeCo. This demonstrates that our method provides more effective conditioning signals, enabling the model to better leverage the existing training data.

\subsection{KoCo Accelerates Model Convergence During Pre-training}

\begin{figure*}[!t]
    \centering
    
    \includegraphics[width=\linewidth]{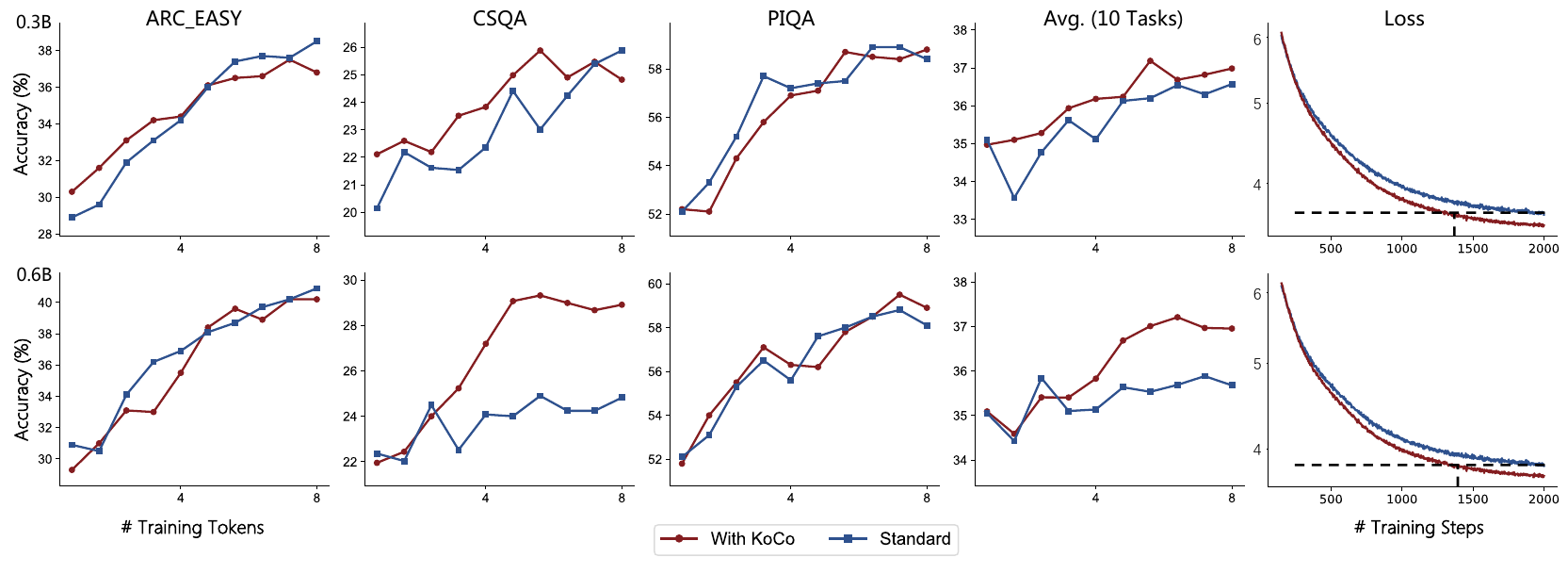}
    \caption{Pre-training from scratch results comparing KoCo (red) with standard paradigm (blue) on 0.3B and 0.6B models. KoCo consistently outperforms standard training across most tasks and achieves lower training loss, demonstrating accelerated convergence.}
    \label{fig:exp-2}
    
\end{figure*}

To further validate the effectiveness of KoCo, we conduct experiments pre-training LLaMA-architectured models from scratch models at 0.3B and 0.6B parameters. We compare KoCo with the standard training paradigm across multiple downstream tasks and analyze the training dynamics. As shown in Figure~\ref{fig:exp-2}, the results demonstrate that KoCo substantially accelerates pre-training across both model scales. For the 0.6B model, KoCo consistently outperforms standard training across the majority of downstream tasks, with particularly notable improvements on tasks like CSQA and the averaged performance across all 10 tasks. The 0.3B model exhibits similar trends, showing that the benefits of knowledge coordinate conditioning are consistent across different model scales.

Besides, the training loss curves reveal that KoCo achieves faster convergence compared to standard training. Both the 0.3B and 0.6B models trained with KoCo reach lower loss values more quickly. 
These results indicate that KoCo enables more data-efficient LLM training, achieving approximately 30\% convergence speedup.

\subsection{Ablation Study}
We set up ablation studies to investigate how each dimension of the knowledge coordinate (Source, Content, Stability) contributes to the overall performance. We sequentially remove each dimension from the knowledge coordinate tag and train models with the remaining two dimensions. At inference time, we use the same evaluation protocol as the main experiments.

\begin{table}[b]
    \centering
    \resizebox{\linewidth}{!}{
        \begin{tabular}{lccccc}
          \toprule
          \textbf{Setting} & \textbf{ARC-e} & \textbf{ARC-c}  &
          \textbf{OBQA}  & \textbf{PIQA} & \textbf{Avg.} \\
          \midrule
          w/o Source & 76.20 & 44.11 & 50.20 & 73.70 & 53.43 \\
          w/o Content & 76.60 & 43.60 & 51.20 & 74.10 & 53.46 \\
          w/o Stability & 76.70 & 43.09 & 51.00 & 73.80 & 53.32 \\
          \midrule
          KoCo (full) & \textbf{77.40} & \textbf{44.11} & \textbf{51.20} & \textbf{74.80} & \textbf{53.48} \\
          \bottomrule
        \end{tabular}
    }
    \caption{Ablation study on the impact of each knowledge coordinate dimension.}
    \label{tab:knowledge_context_tag_elements}
  \end{table}

Table~\ref{tab:knowledge_context_tag_elements} presents the results. We observe that removing any dimension from the knowledge coordinate results in performance degradation, confirming that each dimension captures complementary information beneficial for pre-training. 
This result supports our hypothesis that the three-dimensional knowledge coordinate provides orthogonal and complementary signals.
\section{Analysis}

\subsection{Effectiveness of Conditional Inference}
\label{sec:steering}

In the previous experiments, we evaluated KoCo-trained models using a generic inference setting without any prefix. In this analysis, we explore the ability of KoCo to steer model behavior at inference time by providing task-specific knowledge coordinates.

We hypothesize that different tasks require different types of knowledge support. For instance, commonsense reasoning may benefit from factual knowledge, while social interaction tasks might align better with discussion-oriented content. To evaluate this, we design task-specific prefixes based on the features of each benchmark. For example, we use \textit{\{Source: Media; Content: Discussion\}} for Social IQA to align with social interaction contexts, and \textit{\{Source: Academic; Content: Pedagogical\}} for LogiQA to target structured logical reasoning. For comparison, we also evaluate MeCo using the customized URL prefixes proposed in their paper~\cite{gaometadata}.

\begin{table}[b]
  \centering
  \resizebox{\linewidth}{!}{
      \begin{tabular}{lccccccccccc}
        \toprule
        \textbf{Model} & \textbf{COPA} & \textbf{IFEval} & \textbf{SIQA} & \textbf{LogiQA}    & \textbf{TruQA} & \textbf{Avg.} \\
        \midrule
        MeCo & 82.00 & 20.02 & 52.90 & 25.50 & 36.33 & 52.42 \\
        + cond. prefix & 83.00 & 20.98 & 53.90 & 25.65 & 36.09 & 52.97 \\ 
        \midrule
        KoCo & 83.00 & 25.54 & 53.40 & 26.88 & 36.61 & 53.48 \\
        + cond. prefix & 84.00 & 27.50 & 53.60 & 28.26 & 40.39 & 54.45 \\
        & \greenbg{$\uparrow 1.00$} & \greenbg{$\uparrow 1.96$} & \greenbg{$\uparrow 0.20$} & \greenbg{$\uparrow 1.38$} & \greenbg{$\uparrow 3.78$} & \greenbg{$\uparrow 0.97$} \\
        \bottomrule
      \end{tabular}
  }
  \caption{Performance comparison with task-specific prefixes.}
  \label{tab:task_specific_prefix}
\end{table}

As shown in Table~\ref{tab:task_specific_prefix}, applying task-specific prefixes improves performance for both MeCo and KoCo, demonstrating that conditional inference is an effective strategy for metadata-conditioned models. Notably, KoCo achieves larger gains from this conditioning: while MeCo improves by 0.54\% on average, KoCo achieves a 0.97\% average improvement. TruthfulQA shows the most substantial difference, where KoCo gains 3.78\%. These results suggest that the structured knowledge coordinates in KoCo provide more effective control signals than URL-based conditioning, enabling better alignment between the conditioning prefix and the model's internal knowledge representations.

\subsection{Mitigating Hallucination via Source Awareness}

The previous analysis reveals that KoCo achieves the most substantial improvement on TruthfulQA, a benchmark specifically designed to evaluate model hallucination. TruthfulQA contains questions where models tend to generate false answers due to common misconceptions or imitative falsehoods learned from web data. 

This observation motivates a deeper investigation into how KoCo's source awareness mechanism contributes to hallucination mitigation.
We design a controlled experiment by constructing knowledge coordinate prefixes with varying levels of source reliability. Specifically, we select prefixes representing unreliable sources paired with ephemeral, opinion-based content types. For comparison, we construct prefixes representing reliable sources paired with instructional content types. We then evaluate the model on TruthfulQA using each prefix configuration.

As shown in Table~\ref{tab:model_hallucination}, the model exhibits clear sensitivity to the provided coordinates. When conditioned on prefixes associated with noisy social media sources, the model's truthfulness score drops to around 35\%, below the baseline without any prefix. In contrast, prefixes indicating high reliability boost the score significantly.

This result demonstrates that KoCo learns the correlation between source characteristics and factual reliability during pre-training. More importantly, it provides a practical mechanism for hallucination control: users can suppress unreliable outputs simply by specifying a trustworthy source context at inference time.

\begin{table}[t]
  \centering
  \resizebox{\linewidth}{!}{
      \begin{tabular}{cccc}
        \toprule
        \textbf{Source} & \textbf{Content} & \textbf{Stability}  &
        \textbf{TruQA} \\
        \midrule
        Personal (x.com) & Opinion & Ephemeral & 34.75 \\
        Community (4chan.org) & Opinion & Ephemeral & 35.07 \\
        Community (reddit.com) & Discussion & Ephemeral & 35.35 \\
        \midrule
        - & - & - & 36.61 \\
        Academic (Official Docs) & Instructional & Long-term & 37.89 \\
        Publication (Official FAQ) & Instructional & Long-term & \textbf{40.39} \\
        \bottomrule
      \end{tabular}
  }
  \caption{Impact of different knowledge coordinate prefixes on TruthfulQA. Prefixes associated with unreliable or ephemeral sources (top) degrade truthfulness, while reliable sources (bottom) significantly enhance it.}
  \label{tab:model_hallucination}
\end{table}

\subsection{Internal Mechanisms}

We further explore the internal mechanisms behind KoCo from the perspectives of prediction uncertainty and representation learning.

\textbf{Perplexity.} 
We analyze KoCo from the perspective of prediction uncertainty. Using 1,024 documents unseen during pre-training, we compute the perplexity under two conditions: with the correct knowledge coordinate prefix and with a randomly assigned prefix.
As shown in Table~\ref{tab:perplexity}, providing the correct knowledge coordinate significantly reduces perplexity compared to random coordinates (5.87 vs. 6.04). This confirms that the model has learned meaningful associations between coordinates and document characteristics. The correct coordinate effectively narrows down the prediction space, enabling more accurate next-token predictions.

\begin{table}[ht]
  \centering
  \small
      \begin{tabular}{lc}
        \toprule
        \textbf{Model Setting} & \textbf{Perplexity} \\
        \midrule
        KoCo + random prefix & 6.04 \\
        KoCo + correct prefix & \textbf{5.87} \\
        \bottomrule
      \end{tabular}
  \caption{Perplexity comparison on unseen documents. The correct knowledge coordinate (prefix) significantly reduces perplexity compared to random coordinates.}
  \label{tab:perplexity}
\end{table}

\textbf{Representations.} 
To explore how the model internally distinguishes between different types of knowledge, we select 20 high-stability factual statements and 20 low-stability opinion-based statements, representing two distant categories in our knowledge coordinate space. We extract the final layer hidden states from the last 50\% of tokens and apply PCA for visualization, comparing KoCo with standard pre-trained model.

\begin{figure}[t]
    \centering
    \includegraphics[width=\linewidth]{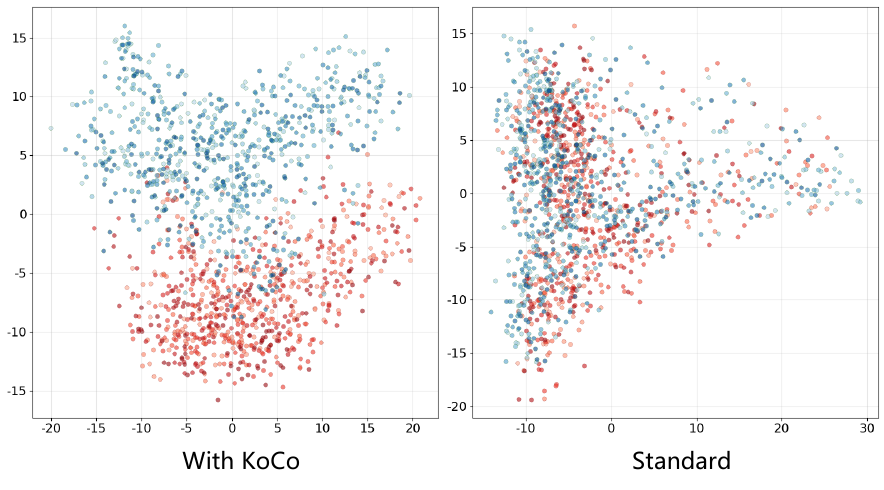}
    \caption{PCA visualization of the hidden states for factual (blue) vs. opinion-based (red) statements. KoCo learns to separate these concepts into distinct clusters, while the standard model shows substantial overlap.}
    \label{fig:representation_visualization}
\end{figure}

As shown in Figure~\ref{fig:representation_visualization}, KoCo exhibits a clear separation between factual and opinion-based statements, with the two categories forming distinct clusters in the embedding space. This separation implies that KoCo learns a disentangled semantic space where different types of knowledge are encoded differently, rather than merely memorizing tags. This disentanglement likely contributes to KoCo's enhanced ability to mitigate hallucinations.

\subsection{Data Distribution and Tagger Analysis}

\begin{figure}[!t]
  \centering
  \includegraphics[width=\linewidth]{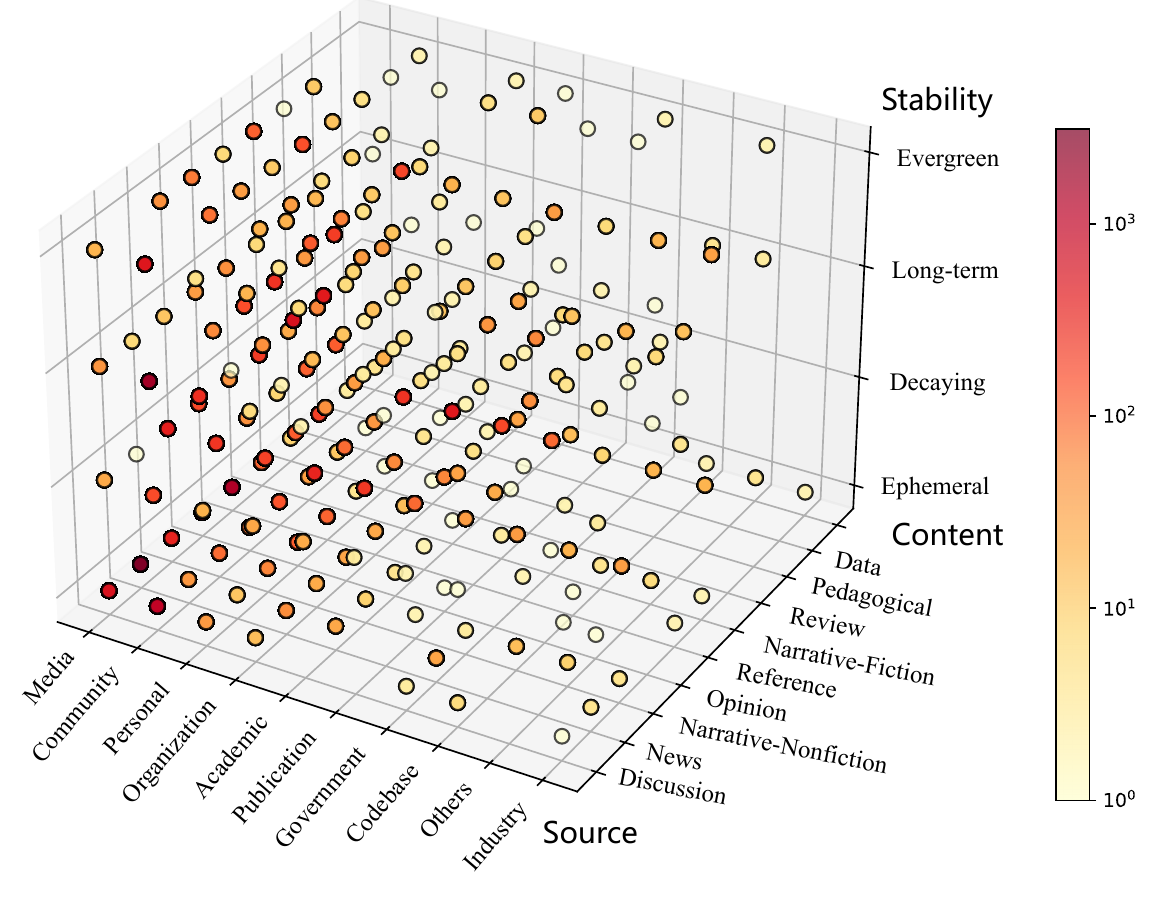}
  \caption{Distribution of 10,000 documents sampled from the DCLM corpus in the knowledge coordinate.}
  \label{fig:dist}
\end{figure}

Figure~\ref{fig:dist} visualizes the distribution of 10,000 randomly sampled DCLM corpus in our knowledge coordinate space. The distribution indicates that documents from Media and Community sources with Discussion or Opinion content types dominate the corpus, while Academic and Government sources are relatively sparse. This imbalance reflects the natural composition of web-scale data. It highlights the importance of coordinate conditioning, without which models may learn patterns from dominant but potentially less reliable sources.

\textbf{Tagger Validation.}
Since our approach relies on an external lightweight model (Qwen3-4B) to predict knowledge coordinates, we validate its reliability by comparing predictions with commercial SOTA LLMs including Gemini-3 and GPT-5.1. On held-out 1,000 samples, the average consistency across the Source, Content and Stability dimensions is 81.8\%, 82.6\%, and 75.5\%, respectively. The moderate agreement suggests that while coordinate classification involves some subjectivity, the overall consistency confirms that our taxonomy captures meaningful distinctions recognized across different LLMs.

\textbf{Does KoCo Distill from Tagger?}
A natural question is whether KoCo's improvements stem from distilling knowledge from the tagger model. To investigate this, we train a lightweight BERT-base (110M parameters) on 50K labeled samples as an alternative tagger, which is notably much smaller than the 0.6B model being pre-trained. We then pre-train KoCo from scratch using coordinates generated by this weaker classifier.

\begin{figure}[!t]
  \centering
  \includegraphics[width=\linewidth]{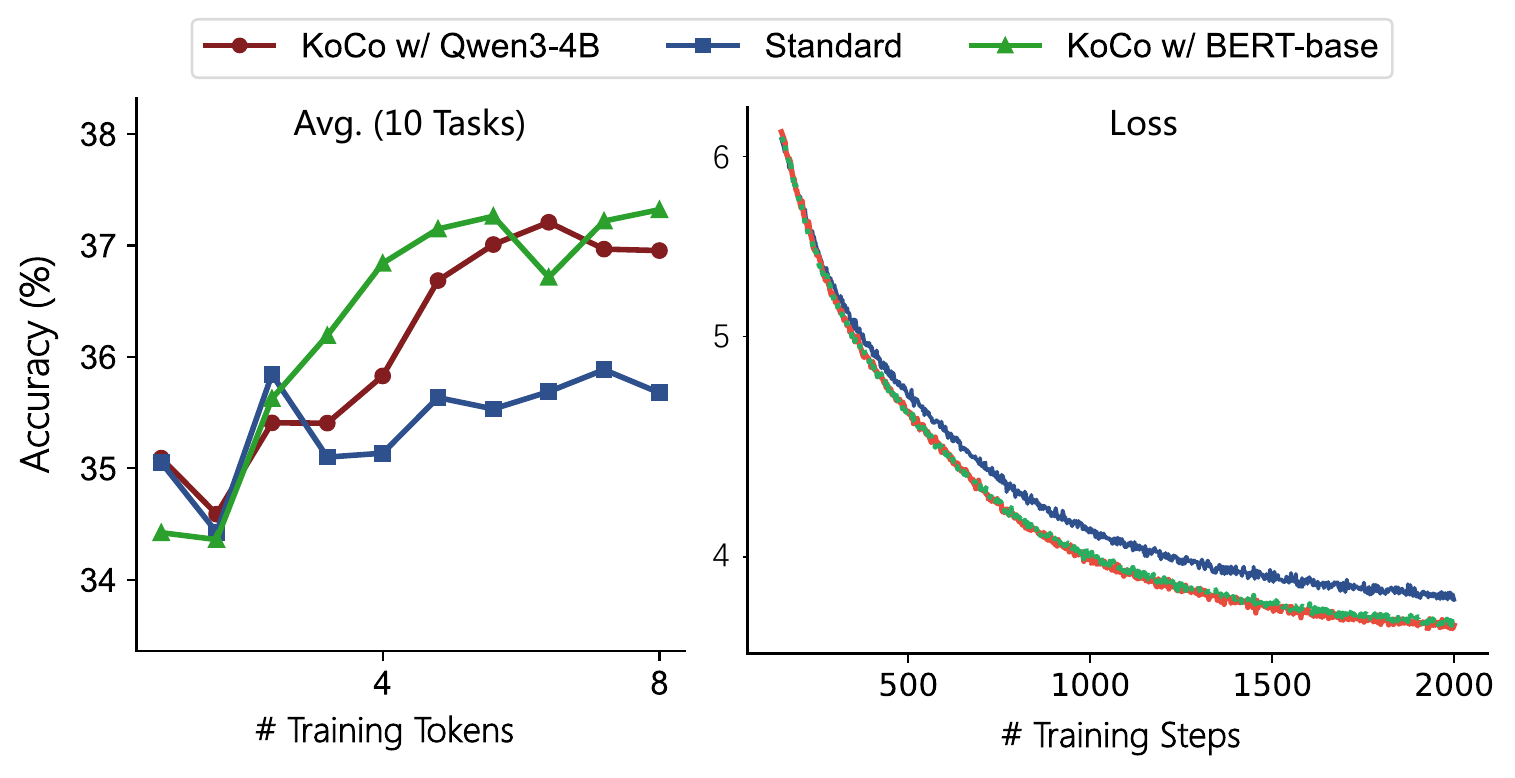}
  \caption{Comparison of KoCo with different taggers. KoCo with BERT-base achieves comparable performance to KoCo with Qwen3-4B, both outperforming standard pre-training.}
  \label{fig:bert-cmp}
\end{figure}

As shown in Figure~\ref{fig:bert-cmp}, KoCo with BERT-base tagger achieves comparable performance to KoCo with Qwen3-4B, and both significantly outperform standard pre-training. Since the BERT-base tagger is much smaller than the pre-trained model, the performance gains are unlikely to result from knowledge distillation. Instead, these results indicate that KoCo benefits from the coordinate conditioning mechanism itself, where the coordinates provide signals to structure knowledge acquisition.

\section{Discussion}

\textbf{Complementarity with Data Selection.}
While data selection methods enhance performance by refining the training distribution, they inevitably trade off data quantity and diversity for quality. KoCo offers an alternative perspective: low-quality data is often just information in the wrong context. Instead of simply discarding data, KoCo enables the model to effectively utilize a broader range of corpora by clarifying their nature. In addition, our approach is complementary to data selection paradigms, it can be integrated with selection pipelines to further structure high-quality data, thereby maximizing both data efficiency and model performance.

\textbf{Role of Context-Aware Pre-training.}
Human cognitive development relies heavily on explicit priors during early learning stages. KoCo provides a similar tool for language models. By supplying knowledge contexts, it offers guidance that is particularly effective during the cold start phase of pre-training. Nevertheless, we recognize that KoCo functions as an accelerator for convergence rather than a substitute for scale. While it improves learning efficiency, the ultimate capabilities of the model remain fundamentally constrained by scaling laws.

\textbf{Bridging Pre-training and Alignment.}
KoCo introduces control signals at the pre-training stage, a mechanism typically reserved for post-training alignment (e.g., SFT or RLHF). This structure allows for intrinsic steerability and blurs the boundary between pre-training and fine-tuning. This implies that certain alignment objectives can be moved upstream to the pre-training phase, potentially simplifying downstream tuning processes. 

\section{Conclusion}

In this paper, we introduced Knowledge Coordinate Conditioning (KoCo), a simple yet effectivepre-training framework that maps heterogeneous web corpora into a three-dimensional semantic space. By explicitly conditioning the model on the Source, Content, and Stability of each document, KoCo mimics the human cognitive process of contextualizing information, addressing the limitations of treating corpora as flattened token sequences.

Experiments show that KoCo outperforms existing baselines, accelerates convergence, and mitigates hallucinations by distinguishing reliable from unreliable sources.
These findings suggest that structuring the wild web data with objective coordinates is a promising direction for building more efficient, robust, and controllable LLMs.

\section*{Limitations}

While our results demonstrate the effectiveness of KoCo, we identify several limitations and avenues for future research.

\paragraph{Scaling to Larger Models.}
Due to computational constraints, our experiments were conducted on models up to 1.6B parameters. While we observed consistent improvements across 0.3B, 0.6B, and 1.6B scales, validating KoCo on larger foundational models (e.g., 7B or larger) trained on multi-trillion tokens remains an important next step. 

\paragraph{Dependency on Proxy Model.}
Our method relies on an upstream model to generate knowledge coordinates. Although our ablation study with a BERT-base tagger suggests that the method is robust to the tagger's capacity to some extent, the quality of the taxonomy and the accuracy of the tagging process fundamentally bound the performance.

\section*{Ethical Statement}

This work utilizes public web data from the DCLM corpus. We acknowledge that web-scale corpora inherently contain biases and sensitive information. A specific ethical consideration for KoCo is the potential bias in the automated tagging process. The upstream tagger may reflect societal biases, potentially misclassifying content from underrepresented groups. We attempt to mitigate this by defining objective coordinate dimensions rather than subjective quality judgments, but we recognize that algorithmic bias in the tagger remains a limitation. Positively, by accelerating pre-training convergence, our method might contribute to reducing the energy footprint of LLM development.

\section*{Acknowledgements}

This work was supported by the National Natural Science Foundation of China under Grant 62576216 , Guangdong Provincial Key Laboratory under Grant 2023B1212060076. We are also grateful to Peiru Yang, Ruiqi Zhou, Xintian Li for helpful ideas and feedback.

\bibliography{custom}
\clearpage

\appendix
\section{Experiment Details}

\subsection{Hyperparameters}
Table~\ref{tab:hyperparameter} shows the hyperparameter settings used in our experiments. 

\begin{table}[H]
  \centering
  \small
      \begin{tabularx}{\linewidth}{lX}
        \toprule
        \textbf{Hyperparameters} & \textbf{Values} \\
        \midrule
        Optimizer & AdamW ($\beta_1 = 0.9$, $\beta_2 = 0.95$) \\
        Learning rate & $3e - 4$ \\
        Weight decay & 0.033 \\
        Batch size & 4M tokens \\
        Warmup & 5\% linear warmup \\
        Schedule & Cosine decay to 10\% of the peak learning rate \\
        Seq length & Pack to 8192 tokens \\
        \bottomrule
      \end{tabularx}
  \caption{Hyperparameter settings for our experiments.}
  \label{tab:hyperparameter}
\end{table}

\subsection{Model configurations}
We use the Llama variant~\cite{touvron2023llama} of Transformers~\cite{vaswani2017attention} for our experiments. All models use the Llama-3 tokenizer~\cite{dubey2024llama}. We add a BOS and an EOS token at the beginning and end of every document. The detailed configurations are specified in Table~\ref{tab:model_configuration}

\begin{table}[H]
  \centering
  \small
      \begin{tabular}{ccccc}
        \toprule
        \textbf{Param} & \textbf{Layers} & \textbf{Inter} & \textbf{Hidden} & \textbf{Heads} \\
        \midrule
        300M & 20 & 3072 & 768  & 12 \\
        600M & 24 & 4096 & 1024 & 16 \\
        1.6B & 24 & 5504 & 2048 & 16 \\
        \bottomrule
      \end{tabular}
  \caption{Model configurations for our experiments.}
  \label{tab:model_configuration}
\end{table}

\subsection{Experimental resource}
Table~\ref{tab:experimental_resource} shows the resources required to train the models in our experiments. Our main models (1.6B, 8B tokens) take roughly 2 days to train on 4 A800 GPUs.

\begin{table}[H]
  \centering
  \small
      \begin{tabular}{cccccc}
        \toprule
       Param & 300M & 600M & 1.6B \\
       Tokens & 8B & 8B & 8B \\
        \midrule
        GPU hours & 60 & 96 & 192 \\
        \bottomrule
      \end{tabular}
  \caption{Resources required to train the models in our experiments (A800 GPU hours).}
  \label{tab:experimental_resource}
\end{table}

\subsection{Prompt for tagging knowledge coordinate}
Table~\ref{tab:prompt_for_tagging} shows the prompt used for tagging knowledge coordinats. We prompt a Qwen3-4B model to tag knowledge coordinate.

\begin{table*}[ht]
  \centering
  \small
      \begin{tabular}{p{16cm}}
        \toprule
\# Role: Knowledge Taxonomist \& Data Curator

\# Core Task: Your task is to analyze the provided [TEXT] and its source [URL], and generate a single, concise ``Knowledge Context Meta-Tag.'' This tag serves solely to inform the training model of the ``position'', ``sourcle'', and ``attribute'' of the text it will read within the broader human knowledge system.

Note that your goal is to **classify text**, not **summarize the content of the text**. Do not expand or modify existing information.

---

\# Step Guide:

1.  **Analysis [URL]:**   

    * Check the domain name (e.g. `wikipedia.org', `nature.com', `615vgs.com', `nytimes.com').
    
    * Infer **Source Type**. Classification includes but is not limited to:
    
        * `Primary Academic' (such as nature.com)    
        
        * `Secondary Academic' (such as textbook websites)   
        
        * ..
        
2. **Analysis [TEXT]:** 

* Determine **Topic Domain** (e.g., basic physics, aerospace history, finance, entertainment).

    * Evaluate **Context** including but not limited to:
    
    * `Core Principle'       
    
    * `Specialized Knowledge'       
    
    * `Factual Report'        
    
    * ...    
    
    * Define **Temporal Stability ($F_S$)** to measure the ability of knowledge to resist semantic or structural changes over time. High $F_S$ indicates that the knowledge has long-term validity, broad consensus, and low temporal fluctuation    
    
    * Evaluate **Temporal Stability (Stability - $F_S$)**:        
    
    * `Evergreen'         
    
    * `Long-term'        
    
    * `Decaying'     
    
    * `Ephemeral' 

3. **Generating meta-labels:

** Combine the above analysis into a **single descriptive sentence**.

    * Use the following **strict format**:
    
    `Source: <Source Type> (<Domain Name>). Content: A paragraph of text describing <Importance> and <Field of Knowledge>. Stability: <Timeliness>'
    
---

\# Example:

**[Example: Newton's Laws]**

* **[TEXT]**: (An excerpt from a textbook describing Newton's second law F=ma)

* **[URL]**: `https://physics.stackexchange.com/questions/123'

* **[Output]**: `Source: Q\&A Forum (physics.stackexchange.com). Content: A definition of a Core Principle within the Basic Physics domain. Stability: Evergreen.'
---

Please generate meta tags for the corpus and URL you provided. Simply output your results in English

**[TEXT]**

\{text\}

**[URL]**

\{url\}
        \\
    \bottomrule
    \end{tabular}
  \caption{The prompt for tagging knowledge coordinate.}
  \label{tab:prompt_for_tagging}
\end{table*}

\subsection{Customized prefixes for conditional inference}
Table~\ref{tab:customized_prefixes} shows the customized prefixes for conditional inference.

\begin{table}[H]
  \centering
  \small
  \resizebox{\linewidth}{!}{
      \begin{tabular}{lccc}
        \toprule
        \textbf{Tasks} & \textbf{Source} & \textbf{Content} & \textbf{Stability} \\
        \midrule
        COPA & KnowledgeBase (stackenfm.com) & Knowledge  & Ephemeral \\
        ARC-Easy & KnowledgeBase(simple.wikipedia.org) & Knowledge & Evergreen \\
        ARC-Challenge & KnowledgeBase(simple.wikipedia.org) & Knowledge & Evergreen \\
        CommonsenseQA & Textbook(textbook.com) & Instructional & Evergreen \\
        IFEval & Official FAQ(faq.com)  & Instructional & Long-term \\
        OpenBookQA & KnowledgeBase(simple.wikipedia.org) & Knowledge & Evergreen  \\
        PIQA & KnowledgeBase(simple.wikipedia.org) & Knowledge & Evergreen  \\
        Social IQA & Official Docs(technic.docs.com) & Instructional & Decaying  \\
        LogiQA & KnowledgeBase(en.wikipedia.org) & Knowledge & Evergreen  \\
        TruthfulQA & Official FAQ(faq.com) & Instructional & Long-term \\
        \bottomrule
      \end{tabular}
  }
  \caption{Customized prefixes for conditional inference}
  \label{tab:customized_prefixes}
\end{table}

\section{Full Results}
Table~\ref{tab:full_results_ablation_study}, Table~\ref{tab:0.3B-parameter-8Btokens} and Table~\ref{tab:0.6B-parameter-8Btokens} show the detailed results of experiments reported in our paper. Table~\ref{tab:0.3B-parameter-24Btokens} presents additional results on extended pre-training with 3$\times$ more training data (24B tokens), demonstrating that KoCo's advantage persists and grows with more training data.

\begin{table*}[h]
  \centering
  \small
  \resizebox{\linewidth}{!}{
      \begin{tabular}{lccccccccccc}
        \toprule
        \textbf{Setting} & \textbf{COPA} & \textbf{ARC-e} & \textbf{ARC-c} & \textbf{CSQA} & \textbf{IFEval} &
        \textbf{OBQA}  & \textbf{PIQA} & \textbf{SIQA} & \textbf{LogiQA}    & \textbf{TruQA} & \textbf{Avg} \\
        \midrule
        w/o Source & 84.00 & 76.20 & 44.11 & 62.08 & 26.38 & 50.20 & 73.70 & 54.00 & 27.34 & 36.32 & 53.43 \\
        w/o Content & 84.00 & 76.60 & 43.60 & 62.49 & 26.38 & 51.20 & 74.10 & 53.10 & 27.04 & 36.10 & 53.46 \\
        w/o Stability & 82.00 & 76.70 & 43.09 & 62.49 & 25.78 & 51.00 & 73.80 & 53.20 & 27.34 & 37.87 & 53.32 \\
        \midrule
        KoCo & 83.00 & 77.40 & 44.11 & 61.83 & 25.54 & 51.20 & 74.80 & 53.40 & 26.88 & 36.61 & 53.48 \\
        \bottomrule
      \end{tabular}
  }
  \caption{Detailed results on ablation study}
  \label{tab:full_results_ablation_study}
\end{table*}

\begin{table*}[h]
  \centering
  \small
  \resizebox{\linewidth}{!}{
      \begin{tabular}{lccccccccccc}
        \toprule
        \textbf{Tokens} & \textbf{COPA} & \textbf{ARC-e} & \textbf{ARC-c} & \textbf{CSQA} & \textbf{IFEval} &
        \textbf{OBQA}  & \textbf{PIQA} & \textbf{SIQA} & \textbf{LogiQA}    & \textbf{TruQA} & \textbf{Avg} \\
        \midrule
        \multicolumn{12}{c}{Standard} \\
        \midrule
        0.8B & 58.00 & 28.90 & 26.62 & 20.15 & 19.90 & 28.80 & 52.10 & 39.40 & 26.42 & 50.73 & 35.10 \\
      1.6B & 48.00 & 29.60 & 23.29 & 22.19 & 17.63 & 27.40 & 53.30 & 38.30 & 24.58 & 51.38 & 33.57 \\
      2.4B & 51.00 & 31.90 & 25.00 & 21.62 & 18.23 & 30.60 & 55.20 & 38.20 & 25.96 & 50.01 & 34.77 \\
      3.2B & 52.00 & 33.10 & 23.38 & 21.54 & 20.38 & 30.00 & 57.70 & 40.30 & 27.34 & 50.45 & 35.62 \\
      4.0B & 51.00 & 34.20 & 24.15 & 22.36 & 22.42 & 27.60 & 57.20 & 39.20 & 24.12 & 48.89 & 35.11 \\
      4.8B & 58.00 & 36.00 & 26.37 & 24.41 & 22.18 & 25.40 & 57.40 & 39.10 & 25.81 & 46.61 & 36.13 \\
      5.6B & 58.00 & 37.40 & 27.82 & 23.01 & 23.26 & 25.80 & 57.50 & 38.50 & 24.73 & 45.98 & 36.20 \\
      6.4B & 59.00 & 37.70 & 26.88 & 24.24 & 22.30 & 27.00 & 58.90 & 39.00 & 26.11 & 44.31 & 36.54 \\
      7.2B & 55.00 & 37.60 & 27.13 & 25.39 & 22.90 & 26.60 & 58.90 & 39.70 & 25.65 & 44.14 & 36.30 \\
      8.0B & 60.00 & 38.50 & 26.37 & 25.88 & 21.94 & 24.60 & 58.40 & 40.80 & 25.81 & 43.42 & 36.57 \\
        \midrule
        \multicolumn{12}{c}{KoCo} \\
        \midrule
        0.8B & 55.00 & 30.30 & 27.73 & 22.11 & 17.63 & 30.20 & 52.20 & 39.60 & 26.42 & 48.51 & 34.97 \\
        1.6B & 53.00 & 31.60 & 25.00 & 22.60 & 23.02 & 30.60 & 52.10 & 38.40 & 24.73 & 49.95 & 35.10 \\
        2.4B & 52.00 & 33.10 & 26.02 & 22.19 & 20.98 & 30.60 & 54.30 & 38.50 & 24.58 & 50.53 & 35.28 \\
        3.2B & 54.00 & 34.20 & 25.60 & 23.51 & 18.23 & 31.40 & 55.80 & 39.50 & 26.42 & 50.68 & 35.93 \\
      4.0B & 51.00 & 34.40 & 26.11 & 23.83 & 23.14 & 31.60 & 56.90 & 37.90 & 25.81 & 51.12 & 36.18 \\
      4.8B & 49.00 & 36.10 & 25.51 & 24.98 & 21.46 & 31.60 & 57.10 & 40.00 & 26.11 & 50.50 & 36.24 \\
      5.6B & 54.00 & 36.50 & 25.60 & 25.88 & 25.42 & 32.80 & 58.70 & 38.50 & 24.88 & 49.58 & 37.19 \\
      6.4B & 51.00 & 36.60 & 25.34 & 24.90 & 24.22 & 33.00 & 58.50 & 38.50 & 26.11 & 48.68 & 36.68 \\
      7.2B & 50.00 & 37.50 & 25.34 & 25.47 & 23.50 & 32.80 & 58.40 & 39.70 & 26.88 & 48.60 & 36.82 \\
      8.0B & 53.00 & 36.80 & 25.51 & 24.82 & 23.50 & 33.20 & 58.80 & 39.20 & 26.73 & 48.28 & 36.98 \\
        \bottomrule
      \end{tabular}
  }
  \caption{Intermediate checkpoint results for the 0.3B-parameter, 8B-token runs.}
  \label{tab:0.3B-parameter-8Btokens}
\end{table*}

\begin{table*}[h]
  \centering
  \small
  \resizebox{\linewidth}{!}{
      \begin{tabular}{lccccccccccc}
        \toprule
        \textbf{Tokens} & \textbf{COPA} & \textbf{ARC-e} & \textbf{ARC-c} & \textbf{CSQA} & \textbf{IFEval} &
        \textbf{OBQA}  & \textbf{PIQA} & \textbf{SIQA} & \textbf{LogiQA}    & \textbf{TruQA} & \textbf{Avg} \\
        \midrule
        \multicolumn{12}{c}{Standard} \\
        \midrule
        0.8B & 54.00 & 30.90 & 24.83 & 22.36 & 17.75 & 32.60 & 52.10 & 39.90 & 25.50 & 50.59 & 35.05 \\
      1.6B & 57.00 & 30.50 & 26.37 & 22.03 & 13.55 & 29.40 & 53.10 & 39.10 & 23.35 & 49.87 & 34.43 \\
      2.4B & 61.00 & 34.10 & 25.34 & 24.49 & 12.11 & 32.80 & 55.30 & 38.50 & 24.58 & 50.20 & 35.84 \\
      3.2B & 58.00 & 36.20 & 25.77 & 22.52 & 5.04 & 31.20 & 56.50 & 38.90 & 25.96 & 50.92 & 35.10 \\
      4.0B & 55.00 & 36.90 & 25.77 & 24.08 & 8.39 & 30.80 & 55.60 & 39.60 & 25.81 & 49.41 & 35.14 \\
      4.8B & 54.00 & 38.10 & 26.37 & 24.00 & 11.75 & 29.00 & 57.60 & 40.30 & 25.19 & 50.04 & 35.64 \\
      5.6B & 54.00 & 38.70 & 25.68 & 24.90 & 11.63 & 28.60 & 58.00 & 39.20 & 25.04 & 49.58 & 35.53 \\
      6.4B & 53.00 & 39.70 & 24.49 & 24.24 & 10.91 & 31.20 & 58.50 & 40.10 & 25.04 & 49.70 & 35.69 \\
      7.2B & 55.00 & 40.20 & 24.66 & 24.24 & 9.95 & 31.20 & 58.80 & 39.60 & 25.65 & 49.55 & 35.88 \\
      8.0B & 53.00 & 40.90 & 24.83 & 24.82 & 9.71 & 30.40 & 58.10 & 40.10 & 25.35 & 49.57 & 35.68 \\
        \midrule
        \multicolumn{12}{c}{KoCo} \\
        \midrule
        0.8B & 58.00 & 29.30 & 27.82 & 21.95 & 19.30 & 28.00 & 51.80 & 39.40 & 25.04 & 50.31 & 35.09 \\
      1.6B & 54.00 & 31.00 & 24.83 & 22.44 & 20.50 & 25.60 & 54.00 & 38.70 & 23.96 & 50.86 & 34.59 \\
      2.4B & 53.00 & 33.10 & 25.77 & 24.00 & 20.50 & 27.40 & 55.50 & 40.00 & 24.12 & 50.69 & 35.41 \\
      3.2B & 51.00 & 33.00 & 25.60 & 25.23 & 21.22 & 25.80 & 57.10 & 39.20 & 25.35 & 50.56 & 35.41 \\
      4.0B & 53.00 & 35.50 & 25.94 & 27.19 & 20.98 & 26.80 & 56.30 & 38.40 & 25.65 & 48.53 & 35.83 \\
      4.8B & 55.00 & 38.40 & 26.62 & 29.07 & 23.02 & 27.80 & 56.20 & 38.80 & 25.96 & 45.98 & 36.68 \\
      5.6B & 56.00 & 39.60 & 25.43 & 29.32 & 23.02 & 29.80 & 57.80 & 38.50 & 25.65 & 44.96 & 37.01 \\
      6.4B & 58.00 & 38.90 & 25.68 & 28.99 & 23.38 & 29.00 & 58.50 & 39.20 & 26.11 & 44.32 & 37.21 \\
      7.2B & 55.00 & 40.20 & 26.02 & 28.67 & 23.26 & 29.80 & 59.50 & 38.70 & 24.88 & 43.64 & 36.97 \\
      8.0B & 56.00 & 40.20 & 24.23 & 28.91 & 22.78 & 30.60 & 58.90 & 39.50 & 24.88 & 43.54 & 36.95 \\
        \bottomrule
      \end{tabular}
  }
  \caption{Intermediate checkpoint results for the 0.6B-parameter, 8B-token runs.}
  \label{tab:0.6B-parameter-8Btokens}
\end{table*}

\begin{table*}[h]
  \centering
  \small
  \resizebox{\linewidth}{!}{
      \begin{tabular}{lccccccccccc}
        \toprule
        \textbf{Tokens} & \textbf{COPA} & \textbf{ARC-e} & \textbf{ARC-c} & \textbf{CSQA} & \textbf{IFEval} &
        \textbf{OBQA}  & \textbf{PIQA} & \textbf{SIQA} & \textbf{LogiQA}    & \textbf{TruQA} & \textbf{Avg} \\
        \midrule
        \multicolumn{12}{c}{Standard} \\
        \midrule
        6B  & 48.00 & 37.60 & 27.13 & 25.80 & 22.54 & 31.20 & 56.90 & 42.00 & 26.42 & 48.61 & 36.62 \\
        12B & 53.00 & 42.10 & 25.85 & 26.95 & 25.66 & 29.40 & 61.20 & 40.40 & 26.73 & 41.95 & 37.32 \\
        18B & 54.00 & 44.70 & 25.68 & 26.54 & 25.30 & 32.40 & 60.90 & 41.20 & 26.88 & 41.08 & 37.87 \\
        24B & 57.00 & 45.20 & 26.28 & 27.27 & 25.06 & 31.80 & 61.90 & 38.90 & 26.42 & 41.63 & 38.15 \\
        \midrule
        \multicolumn{12}{c}{KoCo} \\
        \midrule
        6B  & 58.00 & 39.50 & 25.68 & 29.32 & 22.06 & 28.40 & 60.50 & 38.50 & 25.50 & 41.75 & 36.92 \\
        12B & 61.00 & 44.10 & 26.88 & 39.64 & 21.82 & 32.00 & 62.80 & 40.30 & 26.27 & 38.77 & 39.36 \\
        18B & 64.00 & 48.20 & 28.75 & 39.39 & 24.94 & 31.20 & 63.20 & 41.40 & 26.42 & 38.93 & 40.64 \\
        24B & 67.00 & 49.30 & 27.82 & 38.66 & 22.18 & 34.20 & 62.90 & 39.60 & 27.19 & 39.52 & 40.84 \\
        \bottomrule
      \end{tabular}
  }
  \caption{Intermediate checkpoint results for the 0.3B-parameter extended pre-training runs (up to 24B tokens, 3$\times$ more data than the main experiments). KoCo's advantage over Standard grows from $+0.30$ at 6B tokens to $+2.69$ at 24B tokens.}
  \label{tab:0.3B-parameter-24Btokens}
\end{table*}

\end{document}